\documentclass{article}

    \PassOptionsToPackage{numbers, compress}{natbib}
 \usepackage[preprint]{neurips_2026}

\usepackage[utf8]{inputenc} 
\usepackage[T1]{fontenc}    
\usepackage{hyperref}       
\usepackage{url}            
\usepackage{booktabs}       
\usepackage{amsfonts}       
\usepackage{nicefrac}       
\usepackage{microtype}      
\usepackage{xcolor}         
\usepackage{multicol}
\usepackage{multirow}
\usepackage{makecell}
\usepackage{graphicx}

\usepackage{amsmath}
\usepackage{amssymb}
\usepackage{mathtools}
\usepackage{amsthm}
\usepackage{algorithm}
\usepackage{enumitem}
\setlist{nosep}
\usepackage{subfigure}
\usepackage[table,dvipsnames]{xcolor}

\definecolor{BaseRow}{HTML}{e8e8e8}
\colorlet{PretrainRow}{red!5}
\colorlet{FineRow}{orange!7}
\colorlet{PropertyRow}{yellow!10}
\colorlet{OursRow}{green!10}

\newif\ifclearcmt
\clearcmtfalse   

\ifclearcmt
    \newcommand{\xl}[1]{}
    \newcommand{\yc}[1]{}
    \newcommand{\tw}[1]{}
\else
    \newcommand{\yc}[1]{{\textcolor{orange}{[YC: #1]}}}
    \newcommand{\tw}[1]{{\textcolor{blue}{[TW: #1]}}}
    \newcommand{\xl}[1]{{\textcolor{teal}{[XL: #1]}}}
\fi

\newcommand{\method}{\texttt{AFANet}}

\title{Beyond LLM-Based Reasoning: \\Lightweight GNNs for Agent Failure Attribution}

\author{%
  Ting-Wei Li \\
  University of Illinois \\
  Urbana-Champaign, IL USA \\
  \texttt{twli@illinois.edu} \\
  \And
  Yuanchen Bei \\
  University of Illinois \\
  Urbana-Champaign, IL USA \\
  \texttt{bei4@illinois.edu} \\
  \And
  Xiao Lin \\
  University of Illinois \\
  Urbana-Champaign, IL USA \\
  \texttt{xiaol13@illinois.edu} \\
  \And
  Hanghang Tong \\
  University of Illinois \\
  Urbana-Champaign, IL USA \\
  \texttt{htong@illinois.edu} \\
}

\begin{document}

\maketitle

\begin{abstract}

Large language model (LLM)-based multi-agent systems (MAS) often exhibit complex failure modes, which frequently cause agents to produce incorrect outcomes. This motivates the task of \textit{Agent Failure Attribution (AFA)}: given a failed multi-agent trajectory, identify the faulty agents and their corresponding error types. Existing approaches predominantly rely on LLMs to perform failure attribution, either through direct prompting, fine-tuning on synthetic data or complex agentic pipelines. While effective, these methods incur substantial computational overhead due to long-context processing, expensive post-training and handcrafted workflows. Moreover, empirical evidence shows that even state-of-the-art models achieve limited accuracy on existing benchmarks, suggesting that scaling model size alone is insufficient. In this work, we revisit this task and question the necessity of such expensive generative solutions. We introduce \method, a lightweight graph-based framework that models interaction trajectories through step-level semantic signals and agent-level relationships. We show that with significantly fewer parameters and near-zero inference cost, \method\ (i) matches or outperforms LLM-based baselines, including fine-tuned models on in-domain benchmarks, (ii) maintains robust performance across different GNN architectures and (iii) can be further improved with inexpensive test-time adaptation on the OOD benchmark. Our results suggest that effective agent failure attribution does not require heavy LLM reasoning and a lightweight, structured approach can achieve  strong performance.

\end{abstract}

\begin{figure}[!ht]
  \centering
  \includegraphics[width=1.0\linewidth]{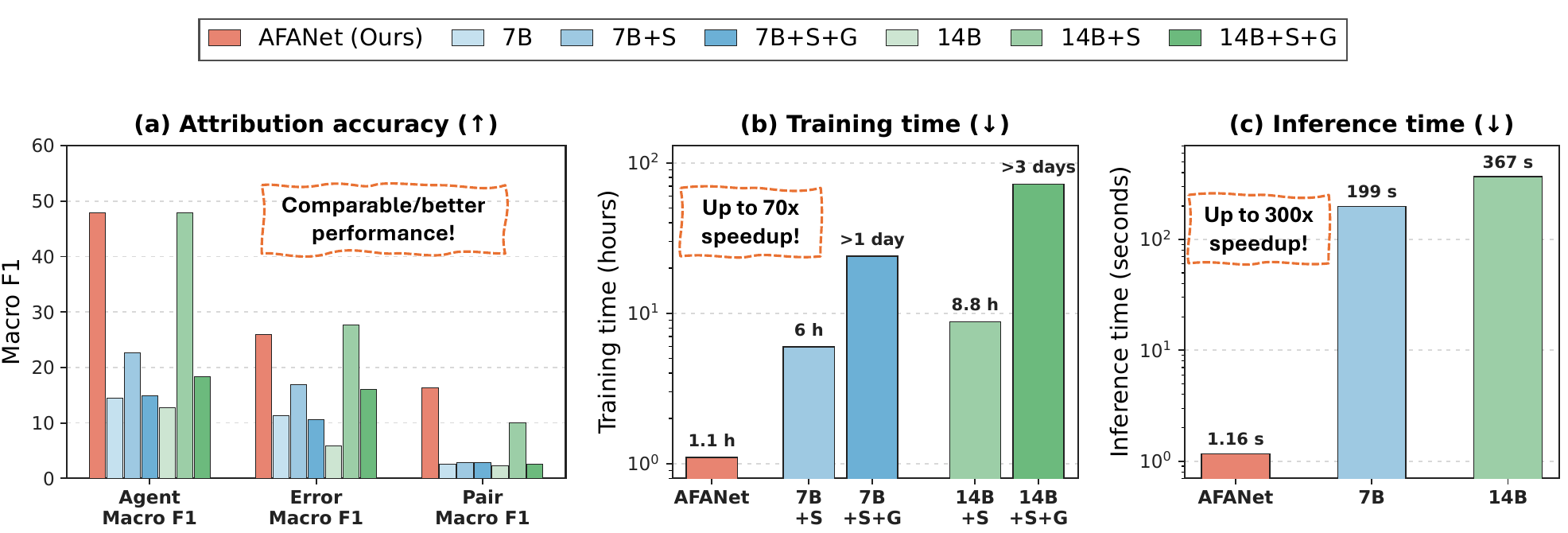}
  \vspace{-5mm}
  \caption{Attribution accuracy and efficiency comparison between our proposed \method\ and LLMs. \method\ can achieve similar or better performance against LLMs while incurring significantly lower training and inference cost. We consider the base models: \texttt{Qwen-2.5-7B/14B-Instruct}, denoted as \textbf{7B}/\textbf{14B}, respectively. We also include their post-trained variants obtained through SFT (\textbf{S}) and subsequent GRPO training (\textbf{G}), denoted by \textbf{7B+S/14B+S} and \textbf{7B+S+G/14B+S+G}.}
  \label{fig:overall-comparison}
\end{figure}

\section{Introduction}
LLM-based multi-agent systems~\citep{li2024survey,guo2024large,wei2026agentic} have emerged as a powerful paradigm for solving complex tasks through coordinated reasoning, tool use, and interaction across multiple agents. Despite their strong capabilities, these systems often exhibit \emph{complex failure modes}~\citep{cemri2025multi,zhang2025agent, wangsurvey, ma2025diagnosing}, where errors introduced by one agent propagate through subsequent interactions and ultimately lead to incorrect outcomes.  This motivates the task of \textit{Agent Failure Attribution (AFA)}~\citep{cemri2025multi,zhang2025agent,deshpande2025trail,kong2025aegis}: given a failed multi-agent interaction trajectory, identify the faulty agents and their corresponding error types responsible for the failure. However, this task is inherently challenging due to long interaction horizons, complex dependencies across agents and the ambiguous nature of error propagation. Even state-of-the-art reasoning models or powerful proprietary models achieve limited accuracy on this task, highlighting the difficulty of identifying root causes within long trajectories~\cite{kong2025aegis}.

Existing approaches predominantly treat failure attribution as a \emph{generative reasoning problem}, relying on LLMs to analyze interaction trajectories and directly generate the answers. These methods typically involve direct prompting~\cite{zhang2025agent}, fine-tuning on synthetic failure data~\cite{kong2025aegis,zhang2025graphtracer,zhang2025agentracer} or complex multi-stage agentic pipelines~\cite{yu2025correct,banerjee2025did,wang2026flat,sun2026scope}. While these approaches have improved performance, they suffer from several fundamental limitations. Firstly \textbf{(Efficiency Bottleneck)}, they incur substantial computational overhead due to long-context processing, repeated inference and expensive training procedures. Secondly \textbf{(Architectural Complexity)}, they introduce significant system complexity through multi-stage workflows and handcrafted pipelines. Lastly \textbf{(Systematical Ineffectiveness)}, despite these efforts, their performance remains limited, sometimes defeated by random baselines, suggesting that scaling model size alone is insufficient for solving this task. These observations raise a fundamental question:
\begin{quote}
\centering
\emph{Is heavy LLM reasoning necessary for agent failure attribution?}
\end{quote}


In this work, we challenge this prevailing paradigm and argue that agent failure attribution can be solved much efficiently with lightweight models. Instead of relying on expensive LLM-based reasoning, we propose \method, a lightweight graph neural network (GNN)~\cite{wu2020comprehensive} that models interaction trajectories through step-level semantic and agent-level structural signals. By explicitly encoding temporal dependencies and inter-agent interactions, \method\ captures the underlying structure of failure propagation. Specifically, \method\ addresses the core challenges of agentic failure attribution in a principled manner. First, graph-based representations naturally model error propagation and inter-agent dependencies, enabling accurate identification of root causes. Second, the model operates with significantly reduced computational cost, avoiding both long-context inference and expensive post-training. Through extensive experiments, we demonstrate that \method\ matches or outperforms strong LLM-based baselines, including supervised fine-tuned and RL-enhanced models, while using significantly fewer parameters and achieving near-zero inference cost (as shown in Figure~\ref{fig:overall-comparison}). We further show that \method\ is robust to architectural choices and can be improved through inexpensive test-time adaptation on the OOD dataset. We summarize our contributions are as follows:
\begin{itemize}


    \item We introduce a new perspective on agent failure attribution: instead of LLM reasoning, graph-based modeling of failure trajectories is sufficient for effective attribution. 
    
    \item We propose \method, a lightweight framework that models multi-agent failure propagation through a turn-level conversation graph, integrating turn-level semantics and temporal/intra-agent dependencies via graph message passing.
    
    
    \item We demonstrate that with substantially lower training and inference cost, \method\ achieves competitive performance compared to LLMs. 
    
    \item We conduct comprehensive ablation study and show that \method\ remains strong under different architectural choices and can be improved via inexpensive test-time adaptation.
\end{itemize}
\section{Preliminaries}
\subsection{Multi-Agent System Trajectories}

Let $\mathcal{A} = \{a_1, a_2, \dots, a_M\}$ denote a set of LLM-based agents.  Given a user query $q \in \mathcal{Q}$, a multi-agent system attempts to solve the task and produces an interaction trajectory $\mathcal{T} = \{(a_{i_t}, x_t)\}_{t=1}^{T}$, where $T$ is the trajectory length, $a_{i_t} \in \mathcal{A}$ is the agent selected at step $t$, and $x_t \in \mathcal{X}$ is the content generated by that agent. 
Each trajectory $\mathcal{T}$ is associated with an outcome variable $o = g(\mathcal{T}, q) \in \{0,1\}$, where $g$ is the outcome verifier and $o = 0/1$ indicates task failure/success, respectively.


\subsection{Agent Failure Attribution (AFA)}

Let $\mathcal{E} = \{e_1, e_2, \dots, e_K\}$ denote a predefined set of error types and $K$ is the number of error types. 
We define the label space as the Cartesian product over agent space and error space: $\mathcal{L} = \mathcal{A} \times \mathcal{E}$. Given a failed trajectory $\mathcal{T}$, the task of \textbf{Agent Failure Attribution} (AFA) is to predict a subset of labels $\mathcal{Y} \subseteq \mathcal{L}$, where each $(a_j, e_k) \in \mathcal{Y}$ indicates that agent $a_j$ commits an error of type $e_k$ that contributes to the failure. We thus formulate AFA as a multi-way classification problem for each agent and our goal is to learn a mapping function $h: \mathcal{T} \rightarrow \{0,1\}^{M \times (K+1)}$  where the first class corresponds to the \textit{clean} state and the remaining
$K$ classes correspond to each error type. 
\section{Methodology}
\begin{figure}[t]
  \centering
  \includegraphics[width=\linewidth]{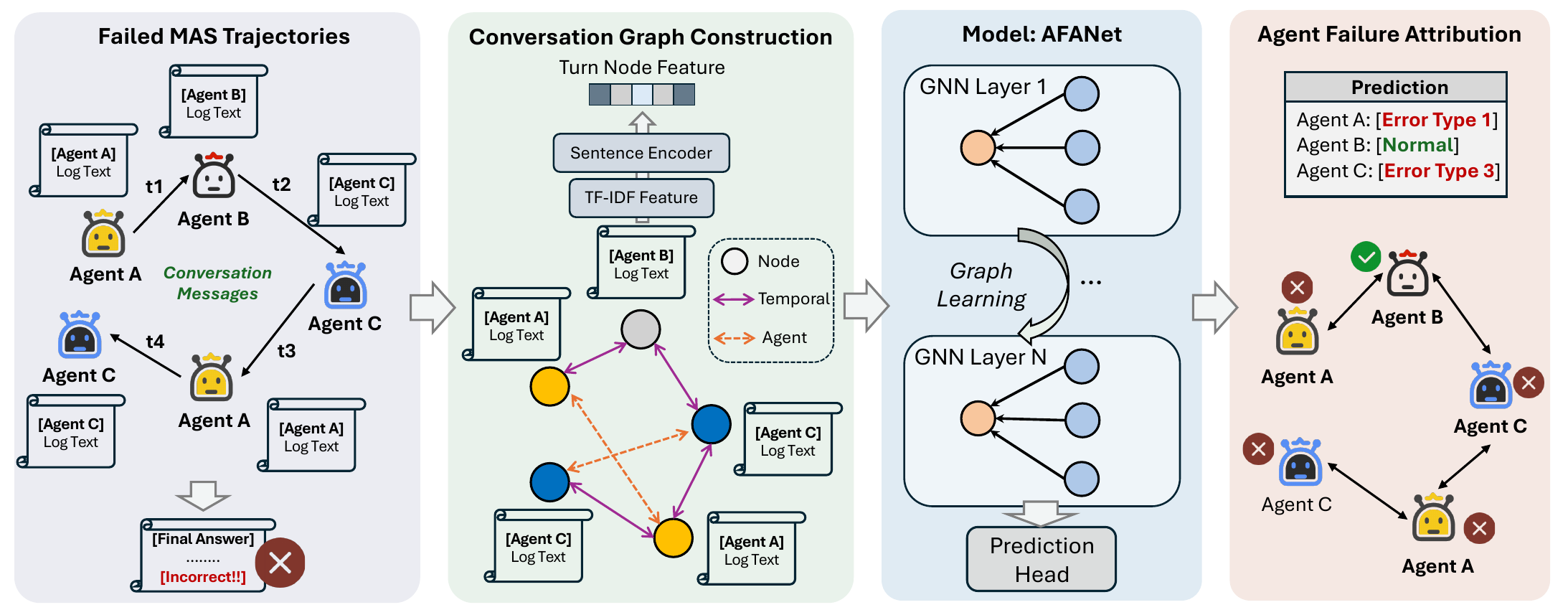}
  \caption{Overall pipeline. Given a failed multi-agent system trajectory, we first transform it into a conversation graph with turn-level textual features and temporal/agent-level connections. The graph will then be passed through our proposed \method\ for agent failure prediction.}
  \label{fig:main-fig}
\end{figure}

In this section, we propose a lightweight graph-based model, \method, to tackle agent failure attribution. Given a conversation, we first construct a turn-level graph with temporal and intra-agent connections (Sec.~\ref{ssec:graph_construction}). We then introduce our proposed model, \method, a simple GNN that combines deviation/statistical signals and contextual semantics to produce agent-level representations (Sec.~\ref{ssec:model_architecture}). Then, we detail the learning objective (Sec.~\ref{ssec:training_objectives}) and finally describe the inference procedure of \method\ (Sec.~\ref{ssec:inference}). The overall pipeline is detailed in Figure~\ref{fig:main-fig}.

\subsection{Conversation Graph Construction}
\label{ssec:graph_construction}
We first transform each multi-agent conversation into a heterogeneous graph with turn-level nodes. The key intuition is that failures in multi-agent systems are not only reflected in the semantic content of individual turns, but also in how an agent's contributions evolve over time and interact with other agents. Therefore, we represent it as a heterogeneous graph whose nodes are conversation turns and whose edges encode temporal and intra-agent dependencies. We detail the construction of node features and edges in the following paragraphs.

\paragraph{Turn node representation.}
Given a failed trajectory $\mathcal{T}=\{(a_{i_t},x_t)\}_{t=1}^{T}$, we construct a graph $\mathcal{G}=(\mathcal{V},\mathcal{E})$,  where each node $v_t \in \mathcal{V}$ corresponds to one turn $(a_{i_t},x_t)$ in the conversation and the number of nodes is $|\mathcal{V}|=T$. 
Inspired by recent anomaly detection literature~\cite{tang2022rethinking,roy2024gad}, we aim to capture abnormal deviation and statistical patterns to derive useful turn-level features for solving AFA. The key intuition is that faulty agent behavior is often reflected through inconsistencies in interaction dynamics rather than only the semantic content of individual turns. 

To obtain such signals efficiently, we first construct per-conversation TF-IDF representations by treating each turn as a document and the entire trajectory as the corpus (more details in Appendix~\ref{app:node-features}). 
We then apply truncated SVD to obtain low-dimensional dense representations: $\mathbf{X}_{\mathrm{svd}} \in \mathbb{R}^{T \times r}$,
where $r$ is the rank size. 
Based on $\mathbf{X}_{\mathrm{svd}}$, we compute a set of deviation  features
that captures how each turn diverges from the contexts in the conversation. We concatenate these deviation features to obtain $\mathbf{x}^{\mathrm{dev}}_t$ for each turn, and stack them across all turns:
$\mathbf{X}_{\mathrm{dev}} = [\mathbf{x}_{\mathrm{dev}}^1| \cdots| \mathbf{x}_{\mathrm{dev}}^T] \in \mathbb{R}^{T\times d_{\mathrm{dev}}}$, where $d_{\mathrm{dev}}$ is the number of deviation features. 

In addition, we consider  statistical features $\mathbf{X}_{\mathrm{stat}} \in \mathbb{R}^{T \times d_{\mathrm{stat}}}$ ($d_{\mathrm{stat}}$ denotes the number of statistical features) that encode non-semantic properties such as positional encoding. Finally, we consider dense semantic features obtained from pretrained sentence encoders (e.g., sentence-BERT~\cite{reimers-2019-sentence-bert}) to capture richer contextual meaning, leading to $\mathbf{X}_{\mathrm{dense}} \in \mathbb{R}^{T \times d_{\mathrm{dense}}}$, where $d_{\mathrm{dense}}$ is the embedding dimension. Stacking these components together, we obtain the initial feature matrix for turn nodes: $
\mathbf{X} = 
\big[
\mathbf{X}_{\mathrm{dev}} \;\big|\; 
\mathbf{X}_{\mathrm{stat}} \;\big|\; 
\mathbf{X}_{\mathrm{dense}}
\big] \in \mathbb{R}^{T \times d}$, where $d = d_{\mathrm{dev}} + d_{\mathrm{stat}} + d_{\mathrm{dense}}$ is the number of feature dimension. Each turn node is therefore associated with a feature vector $\mathbf{x}_t = \mathbf{X}[t,:] \in \mathbb{R}^{d}$. The formal definitions of deviation and statistical features are deferred to Appendix~\ref{app:node-features}.

\paragraph{Edge construction.}
We construct two types of connections: \textit{temporal progression} and \textit{intra-agent dependency}.
Firstly, to preserve the sequential nature of the conversation, we add bidirectional temporal edges between adjacent turns: $(v_t,v_{t+1}) \in \mathcal{E}_{\mathrm{seq}}^{\rightarrow}, (v_{t+1},v_t) \in \mathcal{E}_{\mathrm{seq}}^{\leftarrow}$. Secondly, to capture long-range consistency across non-adjacent turns induced by the same agent, we connect turns produced by the same agent.  For two turns $u<v$ such that $a_{i_u}=a_{i_v}$, we add $(v_u,v_v) \in \mathcal{E}_{\mathrm{agent}}^{\rightarrow}, (v_v,v_u) \in \mathcal{E}_{\mathrm{agent}}^{\leftarrow}$.  Therefore, the full edge set is
$\mathcal{E}
=
\mathcal{E}_{\mathrm{seq}}^{\rightarrow}
\cup
\mathcal{E}_{\mathrm{seq}}^{\leftarrow}
\cup
\mathcal{E}_{\mathrm{agent}}^{\rightarrow}
\cup
\mathcal{E}_{\mathrm{agent}}^{\leftarrow}$. 

\subsection{\method: A Lightweight GNN for Agent Failure Attribution}
\label{ssec:model_architecture}
After constructing the turn-level graph $\mathcal{G}$, we introduce \method, a model that captures both structural and semantic signals to solve AFA. Specifically, turn features are first projected into a hidden space
, followed by graph neural network layers that propagate neighboring turn information along the conversation structure. Finally, agent-level representations are obtained and mapped to prediction scores that quantify the likelihood of each agent being associated with certain error types. We detail the procedure of \method\ as follows.

\paragraph{Input projection.}
We represent each turn node $t$ in the conversation graph with a concatenated feature vector $\mathbf{x}_t  \in \mathbb{R}^{d}$. Before graph message passing, we project these initial node features into a hidden dimension $d_{\mathrm{hidden}}$. This is achieved via a projection module consisting of a linear transformation, a ReLU activation, and layer normalization:
\[
\mathbf{h}_t^{(0)} = \operatorname{LN}(\operatorname{ReLU}(\mathbf{W}_{\mathrm{in}}\mathbf{x}_t + \mathbf{b}_{\mathrm{in}})),
\]
where $\mathbf{W}_{\mathrm{in}}$ is learnable weight matrix, $\mathbf{b}_{\mathrm{in}}$ is learnable bias and $\operatorname{LN}$ denotes layer normalization.

\paragraph{Graph Neural Network blocks.}
To capture contextual dependencies and the structural flow of the conversation, we apply $L$ layers of message passing over the conversation graph. We also incorporate residual skip connections at each layer. The node representation at layer $\ell+1$ is computed as:
\[
\mathbf{h}_t^{(\ell+1)} = \mathbf{h}_t^{(\ell)} + \operatorname{GNN}^{(\ell)}\big( \mathbf{h}_t^{(\ell)}, \{\mathbf{h}_u^{(\ell)} : (u,t)\in\mathcal{E}\} \big),
\]
where $\mathcal{E}$ represents the set of edges in the graph. After $L$ layers, we obtain the per-turn node representation $\mathbf{h}_t^{\mathrm{gnn}} = \mathbf{h}_t^{(L)} \in \mathbb{R}^{d_{\mathrm{hidden}}}$. Note that the $\operatorname{GNN}$ layer can be instantiated with any architecture (e.g. GCN~\cite{kipf2016semi}, GAT~\cite{velivckovic2017graph} and GraphSAGE~\cite{hamilton2017inductive}).

\paragraph{Agent-level pooling.}
Since the final classification objective is performed at the agent level, we aggregate the turn-level representations belonging to each respective agent. For a given agent $a_j$ with the corresponding turn index set $\mathcal{I}_j$, we apply a dual pooling strategy that concatenates both the mean-pooled and max-pooled representations for better expressiveness:
\[
\mathbf{z}_j = \left[ \frac{1}{|\mathcal{I}_j|} \sum_{t \in \mathcal{I}_j} \mathbf{h}_t^{\mathrm{gnn}} \;\Vert\; \max_{t \in \mathcal{I}_j} \mathbf{h}_t^{\mathrm{gnn}} \right].
\]

\paragraph{Prediction head.}
Given the pooled agent representation $\mathbf{z}_j \in \mathbb{R}^{2d_{\mathrm{hidden}}}$, we map it to the prediction space through a multi-layer projection head. We first obtain an intermediate hidden representation for each agent:
\[
\mathbf{h}_j = \operatorname{LN}\!\left(\operatorname{ReLU}(\mathbf{W}_1 \mathbf{z}_j + \mathbf{b}_1)\right),
\]
where $\mathbf{h}_j \in \mathbb{R}^{d_{\mathrm{hidden}}}$. We then apply a bottleneck projection followed by a non-linear activation and dropout:
\[
\tilde{\mathbf{h}}_j = \operatorname{Dropout}\!\left(
\operatorname{ReLU}(\mathbf{W}_{\mathrm{mid}} \mathbf{h}_j + \mathbf{b}_{\mathrm{mid}})
\right),
\]
where $\tilde{\mathbf{h}}_j \in \mathbb{R}^{d_{\mathrm{mid}}}$ and $d_{\mathrm{mid}}$ is the bottleneck dimension size. Finally, we map to the prediction logits:
\[
\mathbf{s}_j = \mathbf{W}_{\mathrm{out}} \tilde{\mathbf{h}}_j + \mathbf{b}_{\mathrm{out}},
\]
where $\mathbf{s}_j \in \mathbb{R}^{K+1}$ represents the un-normalized scores over $K$ classes. The $K+1$ classes correspond to class $0$ for a clean type (i.e. no error occurs) and classes $1,\dots,K-1$ for different error types.

\subsection{Training Objectives}
\label{ssec:training_objectives}

Given the agent-level representations $\{\mathbf{h}_j\}_{j=1}^{M}$, we optimize \method\ using a two-term objective that combines (i) agent-level fault detection and (ii) error-type classification.

\paragraph{Agent-level objective.}
We first derive a binary fault logit $a_j$  from the prediction scores  $\mathbf{s}_j \in \mathbb{R}^{K+1}$ for each agent $j$:
\[
a_j
=
\log \sum_{k=1}^{K} \exp(s_{j,k})
-
\log \exp(s_{j,0}).
\] We then apply a weighted binary cross-entropy loss with logits: 
\[
\mathcal{L}_{\mathrm{agent}}
=
\frac{1}{M}
\sum_{j=1}^{M}
\operatorname{BCEWithLogits}(a_j, y_j^{\mathrm{binary}}; \, w^{+}),
\]
where $y_j^{\mathrm{binary}}=\mathbb{I}[y_j \neq 0]$ denotes whether agent $j$ is faulty and the positive-class weight is defined  as $w^{+} = \frac{n_{\text{clean}}}{n_{\text{faulty}}}$, where $n_{\text{clean}}, n_{\text{faulty}}$ is the number of non-faulty and faulty agents summing over the conversations in the batch. 

\paragraph{Error-level objective.}
For faulty agents $\mathcal{F}=\{j \mid y_j \neq 0\}$, we apply a multi-class cross-entropy loss over the error-type subspace:
\[
\mathcal{L}_{\mathrm{fm}}
=
\frac{1}{|\mathcal{F}|}
\sum_{j \in \mathcal{F}}
\operatorname{CE}([s_{j,1}, \dots, s_{j,K}], y_j - 1).
\]

\paragraph{Overall objective.}
The final loss is simply the weighted sum over the above two terms:
$\mathcal{L}
=
\lambda_{\mathrm{agent}} \mathcal{L}_{\mathrm{agent}}
+
\lambda_{\mathrm{fm}} \mathcal{L}_{\mathrm{fm}}$, where $\lambda_{\mathrm{agent}}, \lambda_{\mathrm{fm}} $ are hyper-parameters.

\subsection{Inference with \method}
\label{ssec:inference}

We first compute the score vector $\mathbf{s}_j \in \mathbb{R}^{K+1}$ for each agent, where $s_{j,0}$ denotes the \emph{clean} class. We derive an agent-level fault score: for each agent $j$,   $\text{score}^{\mathrm{agent}}_j
= \mathbb{P} [\mathrm{agent_j \text{\,is faulty}}] = \frac{\sum_{k=1}^{K} \exp(s_{j,k})}{\sum_{k=0}^{K} \exp(s_{j,k})}$.  During inference, we consider both threshold-based and ranking-based decoding. For \textit{threshold-based decoding}, we utilize  the validation set to sweep the best threshold $\tau$ that maximizes agent-level metric, and then fix $\tau$ to select $\tau_{\mathrm{fm}}$ that maximizes error-level metric. We predict faulty agents as $\hat{\mathcal{F}} = \{ j \mid \text{score}^{\mathrm{agent}}_j \ge \tau \}$. 
For each $j \in \hat{\mathcal{F}}$, we assign $\hat{e}_j = \arg\max_{k \ge 1} s_{j,k}$ and retain $(a_j, \hat{e}_j)$ only if $s_{j,\hat{e}_j} \ge \tau_{\mathrm{fm}}$. For \textit{ranking-based decoding}, we rank agents by $\mathrm{score}^{\mathrm{agent}}_j$ and directly select the top-ranked agent(s), assigning each selected agent the error type with the largest score.

\section{Experiment}


\begin{table*}[t]
\centering
\setlength{\tabcolsep}{4pt}
\renewcommand{\arraystretch}{1.1}
\caption{Performance comparison on AEGIS-Bench and Who\&When. Best and second-best results are in \textbf{bold} and \underline{underline}. All scores are percentages (\%). For the Qwen-based models, \texttt{Instruct} are abbreviated as \textbf{It}, while \texttt{Non-Thinking} and \texttt{Thinking} modes are abbreviated as \textbf{NT} and \textbf{T}. Their SFT and subsequent GRPO-trained variants are denoted by \textbf{+S} and \textbf{+S+G}. Rows marked with $^{**}$ indicate models re-implemented and evaluated by us under the same setting, while the remaining rows are adapted from~\cite{kong2025aegis}. 
} 
\resizebox{\textwidth}{!}{
\begin{tabular}{clccccccccccccc}
\toprule
\midrule
\multirow{4}{*}{\textbf{Category}} 
& \multicolumn{1}{c}{\multirow{4}{*}{\textbf{Method}}}
& \multicolumn{6}{c}{\textbf{AEGIS-Bench}} 
& \multicolumn{6}{c}{\textbf{Who\&When}} 
& \multirow{4}{*}{\textbf{Avg.}} \\
\cmidrule(lr){3-8} \cmidrule(lr){9-14}
 & & \multicolumn{2}{c}{Agent} & \multicolumn{2}{c}{Error} &  \multicolumn{2}{c}{Pair} & \multicolumn{2}{c}{Agent} & \multicolumn{2}{c}{Error} &  \multicolumn{2}{c}{Pair}\\
\cmidrule(lr){3-4} \cmidrule(lr){5-6} \cmidrule(lr){7-8} \cmidrule(lr){9-10}
\cmidrule(lr){11-12} \cmidrule(lr){13-14}
& & $\mu$F1 & MF1 & $\mu$F1 & MF1 & $\mu$F1 & MF1
& $\mu$F1 & MF1 & $\mu$F1 & MF1 & $\mu$F1 & MF1 & \\
\midrule

\textbf{Baseline}
& \cellcolor{BaseRow}Random 
& \cellcolor{BaseRow} 4.54
& \cellcolor{BaseRow} 3.56
& \cellcolor{BaseRow} 11.23
& \cellcolor{BaseRow} 11.15
& \cellcolor{BaseRow} 0.33
& \cellcolor{BaseRow} 0.21
& \cellcolor{BaseRow} 1.06 
& \cellcolor{BaseRow}0.83 
& \cellcolor{BaseRow}8.74 
& \cellcolor{BaseRow}7.14 
& \cellcolor{BaseRow}0.11 
& \cellcolor{BaseRow}0.05 
& \cellcolor{BaseRow}4.08 \\

\midrule

\multirow{6}{*}{\makecell{\textbf{Pre-trained}\\ \textbf{LLMs}}}
& \cellcolor{PretrainRow}Qwen2.5-7B-It 
& \cellcolor{PretrainRow}27.55 & \cellcolor{PretrainRow}14.49
& \cellcolor{PretrainRow}14.96 & \cellcolor{PretrainRow}11.36
& \cellcolor{PretrainRow}5.02 & \cellcolor{PretrainRow}2.52
& \cellcolor{PretrainRow}40.92 & \cellcolor{PretrainRow}23.50
& \cellcolor{PretrainRow}3.64 & \cellcolor{PretrainRow}1.77
& \cellcolor{PretrainRow}2.31 & \cellcolor{PretrainRow}1.14
& \cellcolor{PretrainRow}12.43 \\

& \cellcolor{PretrainRow}Qwen2.5-7B-It$^{**}$ 
& \cellcolor{PretrainRow}26.38 & \cellcolor{PretrainRow}16.15
& \cellcolor{PretrainRow}15.14 & \cellcolor{PretrainRow}11.54
& \cellcolor{PretrainRow}3.67 & \cellcolor{PretrainRow}1.76
& \cellcolor{PretrainRow}46.25 & \cellcolor{PretrainRow}37.85
& \cellcolor{PretrainRow}3.26 & \cellcolor{PretrainRow}1.94
& \cellcolor{PretrainRow}2.06 & \cellcolor{PretrainRow}2.44
& \cellcolor{PretrainRow}14.04 \\

& \cellcolor{PretrainRow}Qwen2.5-14B-It 
& \cellcolor{PretrainRow}35.78 & \cellcolor{PretrainRow}12.71
& \cellcolor{PretrainRow}20.24 & \cellcolor{PretrainRow}5.91
& \cellcolor{PretrainRow}5.47 & \cellcolor{PretrainRow}2.20
& \cellcolor{PretrainRow}49.88 & \cellcolor{PretrainRow}33.19
& \cellcolor{PretrainRow}1.56 & \cellcolor{PretrainRow}1.35
& \cellcolor{PretrainRow}0.00 & \cellcolor{PretrainRow}0.00
& \cellcolor{PretrainRow}13.99 \\

& \cellcolor{PretrainRow}Qwen2.5-14B-It$^{**}$ 
& \cellcolor{PretrainRow}36.72 & \cellcolor{PretrainRow}19.62
& \cellcolor{PretrainRow}18.59 & \cellcolor{PretrainRow}16.49
& \cellcolor{PretrainRow}5.86 & \cellcolor{PretrainRow}2.61
& \cellcolor{PretrainRow}45.47 & \cellcolor{PretrainRow}44.83
& \cellcolor{PretrainRow}0.00 & \cellcolor{PretrainRow}0.00
& \cellcolor{PretrainRow}0.00 & \cellcolor{PretrainRow}0.00
& \cellcolor{PretrainRow}15.85 \\

& \cellcolor{PretrainRow}Qwen3-8B-NT 
& \cellcolor{PretrainRow}21.34 & \cellcolor{PretrainRow}8.16
& \cellcolor{PretrainRow}15.81 & \cellcolor{PretrainRow}13.89
& \cellcolor{PretrainRow}3.96 & \cellcolor{PretrainRow}1.40
& \cellcolor{PretrainRow}27.78 & \cellcolor{PretrainRow}17.64
& \cellcolor{PretrainRow}3.88 & \cellcolor{PretrainRow}1.91
& \cellcolor{PretrainRow}3.88 & \cellcolor{PretrainRow}1.81
& \cellcolor{PretrainRow}10.12 \\

& \cellcolor{PretrainRow}Qwen3-8B-T 
& \cellcolor{PretrainRow}34.63 & \cellcolor{PretrainRow}9.01
& \cellcolor{PretrainRow}17.48 & \cellcolor{PretrainRow}14.31
& \cellcolor{PretrainRow}4.42 & \cellcolor{PretrainRow}1.52
& \cellcolor{PretrainRow}37.91 & \cellcolor{PretrainRow}27.58
& \cellcolor{PretrainRow}4.65 & \cellcolor{PretrainRow}2.21
& \cellcolor{PretrainRow}1.95 & \cellcolor{PretrainRow}1.10
& \cellcolor{PretrainRow}13.06 \\

\midrule

\multirow{9}{*}{\makecell{\textbf{Fine-tuned}\\ \textbf{LLMs}}}
& \cellcolor{FineRow}Qwen2.5-7B-It+S 
& \cellcolor{FineRow}60.03 & \cellcolor{FineRow}22.70
& \cellcolor{FineRow}19.61 & \cellcolor{FineRow}16.90
& \cellcolor{FineRow}5.05 & \cellcolor{FineRow}2.80
& \cellcolor{FineRow}43.51 & \cellcolor{FineRow}32.51
& \cellcolor{FineRow}6.77 & \cellcolor{FineRow}4.20
& \cellcolor{FineRow}1.26 & \cellcolor{FineRow}0.52
& \cellcolor{FineRow}17.99 \\

& \cellcolor{FineRow}Qwen2.5-7B-It+S$^{**}$ 
& \cellcolor{FineRow}37.93 & \cellcolor{FineRow}22.14
& \cellcolor{FineRow}17.69 & \cellcolor{FineRow}13.85
& \cellcolor{FineRow}4.48 & \cellcolor{FineRow}2.02
& \cellcolor{FineRow}\textbf{56.77} & \cellcolor{FineRow}\textbf{65.81}
& \cellcolor{FineRow}5.42& \cellcolor{FineRow}4.75
& \cellcolor{FineRow}3.94 & \cellcolor{FineRow}\textbf{4.58}
& \cellcolor{FineRow}19.97 \\

& \cellcolor{FineRow}Qwen2.5-7B-It+S+G 
& \cellcolor{FineRow}35.43 & \cellcolor{FineRow}14.86
& \cellcolor{FineRow}17.21 & \cellcolor{FineRow}10.54
& \cellcolor{FineRow}7.11 & \cellcolor{FineRow}2.77
& \cellcolor{FineRow}50.77 & \cellcolor{FineRow}30.14
& \cellcolor{FineRow}3.86 & \cellcolor{FineRow}2.30
& \cellcolor{FineRow}2.31 & \cellcolor{FineRow}1.19
& \cellcolor{FineRow}14.87 \\

& \cellcolor{FineRow}Qwen2.5-14B-It+S 
& \cellcolor{FineRow}\textbf{76.53} & \cellcolor{FineRow}\textbf{47.97}
& \cellcolor{FineRow}\textbf{27.53} & \cellcolor{FineRow}\textbf{27.66}
& \cellcolor{FineRow}\underline{16.62} & \cellcolor{FineRow}\underline{9.99}
& \cellcolor{FineRow}51.14 & \cellcolor{FineRow}36.94
& \cellcolor{FineRow}9.87 & \cellcolor{FineRow}{7.77}
& \cellcolor{FineRow}4.03 & \cellcolor{FineRow}2.08
& \cellcolor{FineRow}\textbf{26.51} \\

& \cellcolor{FineRow}Qwen2.5-14B-It+S$^{**}$ 
& \cellcolor{FineRow}41.04 & \cellcolor{FineRow}21.80
& \cellcolor{FineRow}17.16 & \cellcolor{FineRow}13.83
& \cellcolor{FineRow}4.38 & \cellcolor{FineRow}2.02
& \cellcolor{FineRow}49.37 & \cellcolor{FineRow}\underline{64.90}
& \cellcolor{FineRow}1.77& \cellcolor{FineRow}1.73
& \cellcolor{FineRow}0.41 & \cellcolor{FineRow}0.25
& \cellcolor{FineRow}18.22 \\

& \cellcolor{FineRow}Qwen2.5-14B-It+S+G 
& \cellcolor{FineRow}49.74 & \cellcolor{FineRow}18.38
& \cellcolor{FineRow}21.19 & \cellcolor{FineRow}16.10
& \cellcolor{FineRow}6.84 & \cellcolor{FineRow}2.55
& \cellcolor{FineRow}54.43 & \cellcolor{FineRow}40.88
& \cellcolor{FineRow}4.15 & \cellcolor{FineRow}2.67
& \cellcolor{FineRow}2.45 & \cellcolor{FineRow}1.49
& \cellcolor{FineRow}18.41 \\

& \cellcolor{FineRow}Qwen3-8B-NT+S 
& \cellcolor{FineRow}64.79 & \cellcolor{FineRow}38.96
& \cellcolor{FineRow}20.37 & \cellcolor{FineRow}20.36
& \cellcolor{FineRow}9.68 & \cellcolor{FineRow}5.73
& \cellcolor{FineRow}45.48 & \cellcolor{FineRow}30.77
& \cellcolor{FineRow}8.00 & \cellcolor{FineRow}5.29
& \cellcolor{FineRow}5.17 & \cellcolor{FineRow}2.33
& \cellcolor{FineRow}21.41 \\

& \cellcolor{FineRow}Qwen3-8B-NT+S+G 
& \cellcolor{FineRow}45.91 & \cellcolor{FineRow}17.39
& \cellcolor{FineRow}20.89 & \cellcolor{FineRow}15.15
& \cellcolor{FineRow}6.94 & \cellcolor{FineRow}2.82
& \cellcolor{FineRow}50.94 & \cellcolor{FineRow}38.26
& \cellcolor{FineRow}2.21 & \cellcolor{FineRow}1.68
& \cellcolor{FineRow}2.21 & \cellcolor{FineRow}1.45
& \cellcolor{FineRow}17.15 \\

& \cellcolor{FineRow}Qwen3-8B-T+G
& \cellcolor{FineRow}36.11 & \cellcolor{FineRow}15.73
& \cellcolor{FineRow}17.94 & \cellcolor{FineRow}12.03
& \cellcolor{FineRow}4.41 & \cellcolor{FineRow}1.66
& \cellcolor{FineRow}53.12 & \cellcolor{FineRow}40.52
& \cellcolor{FineRow}11.25 & \cellcolor{FineRow}6.91
& \cellcolor{FineRow}\textbf{8.10} & \cellcolor{FineRow}3.19
& \cellcolor{FineRow}17.58 \\

\midrule

\multirow{6}{*}{\makecell{\textbf{Proprietary}\\ \textbf{LLMs}}}
& \cellcolor{PropertyRow}GPT-4.1 
& \cellcolor{PropertyRow}37.48 & \cellcolor{PropertyRow}11.12
& \cellcolor{PropertyRow}20.65 & \cellcolor{PropertyRow}15.75
& \cellcolor{PropertyRow}7.44 & \cellcolor{PropertyRow}2.27
& \cellcolor{PropertyRow}42.29 & \cellcolor{PropertyRow}28.93
& \cellcolor{PropertyRow}7.00 & \cellcolor{PropertyRow}5.84
& \cellcolor{PropertyRow}3.36 & \cellcolor{PropertyRow}1.16
& \cellcolor{PropertyRow}15.27 \\

& \cellcolor{PropertyRow}GPT-4o-mini 
& \cellcolor{PropertyRow}38.54 & \cellcolor{PropertyRow}14.72
& \cellcolor{PropertyRow}19.95 & \cellcolor{PropertyRow}16.02
& \cellcolor{PropertyRow}5.76 & \cellcolor{PropertyRow}1.63
& \cellcolor{PropertyRow}47.42 & \cellcolor{PropertyRow}34.21
& \cellcolor{PropertyRow}5.26 & \cellcolor{PropertyRow}3.33
& \cellcolor{PropertyRow}2.11 & \cellcolor{PropertyRow}0.98
& \cellcolor{PropertyRow}15.83 \\

& \cellcolor{PropertyRow}o3  
& \cellcolor{PropertyRow}40.31 & \cellcolor{PropertyRow}23.27
& \cellcolor{PropertyRow}22.37 & \cellcolor{PropertyRow}16.76
& \cellcolor{PropertyRow}7.86 & \cellcolor{PropertyRow}2.27
& \cellcolor{PropertyRow}53.10 & \cellcolor{PropertyRow}42.55
& \cellcolor{PropertyRow}\textbf{14.88} & \cellcolor{PropertyRow}\underline{8.63}
& \cellcolor{PropertyRow}\underline{7.41} & \cellcolor{PropertyRow}3.98
& \cellcolor{PropertyRow}20.24 \\

& \cellcolor{PropertyRow}Gemini-2.5-Flash 
& \cellcolor{PropertyRow}42.02 & \cellcolor{PropertyRow}16.45
& \cellcolor{PropertyRow}23.47 & \cellcolor{PropertyRow}19.85
& \cellcolor{PropertyRow}6.99 & \cellcolor{PropertyRow}2.76
& \cellcolor{PropertyRow}\underline{55.56} & \cellcolor{PropertyRow}36.98
& \cellcolor{PropertyRow}11.94 & \cellcolor{PropertyRow}7.96
& \cellcolor{PropertyRow}7.32 & \cellcolor{PropertyRow}3.33
& \cellcolor{PropertyRow}19.55 \\

& \cellcolor{PropertyRow}Gemini-2.5-Pro 
& \cellcolor{PropertyRow}41.32 & \cellcolor{PropertyRow}16.15
& \cellcolor{PropertyRow}19.93 & \cellcolor{PropertyRow}16.29
& \cellcolor{PropertyRow}6.96 & \cellcolor{PropertyRow}2.88
& \cellcolor{PropertyRow}53.11 & \cellcolor{PropertyRow}34.92
& \cellcolor{PropertyRow}11.07 & \cellcolor{PropertyRow}{8.11}
& \cellcolor{PropertyRow}6.81 & \cellcolor{PropertyRow}2.69
& \cellcolor{PropertyRow}18.35 \\

& \cellcolor{PropertyRow}Claude-Sonnet-4 
& \cellcolor{PropertyRow}40.73 & \cellcolor{PropertyRow}15.51
& \cellcolor{PropertyRow}21.21 & \cellcolor{PropertyRow}16.55
& \cellcolor{PropertyRow}7.68 & \cellcolor{PropertyRow}2.34
& \cellcolor{PropertyRow}44.76 & \cellcolor{PropertyRow}37.23
& \cellcolor{PropertyRow}13.33 & \cellcolor{PropertyRow}\textbf{9.23}
& \cellcolor{PropertyRow}6.77 & \cellcolor{PropertyRow}2.66
& \cellcolor{PropertyRow}18.16 \\

\midrule

\multirow{1}{*}{\textbf{Ours}}
& \cellcolor{OursRow}\method\ & \cellcolor{OursRow}\underline{74.16} & \cellcolor{OursRow}\underline{47.86} & \cellcolor{OursRow}\underline{27.01} & \cellcolor{OursRow}\underline{25.96} & \cellcolor{OursRow}\textbf{17.42} & \cellcolor{OursRow}\textbf{16.35} & \cellcolor{OursRow}37.93 & \cellcolor{OursRow}20.26 & \cellcolor{OursRow}\underline{13.79} & \cellcolor{OursRow}6.05 & \cellcolor{OursRow}6.90 & \cellcolor{OursRow}\underline{4.16} & \cellcolor{OursRow}\underline{24.82} \\
\midrule
\bottomrule
\end{tabular}
}

\label{tab:main_tab}
\end{table*}

In this section, we conduct extensive experiments to evaluate the effectiveness of \method\ against LLM-based baselines. Our experiments are
designed to answer the following research questions:

\begin{itemize}
    \item (\textbf{RQ1}): How does \method\ compare to LLM-based methods w.r.t. efficacy and efficiency?
    \item (\textbf{RQ2}): How does \method\ perform under different GNN backbones and hyper-parameters?
    \item (\textbf{RQ3}): How does \method\ perform across domains and how to improve it?
\end{itemize}


\subsection{Experimental Setup}
\label{ssec:exp-setup}
\paragraph{Datasets.}

We conduct experiments on the AEGIS-Bench~\cite{kong2025aegis}\footnote{\url{https://huggingface.co/datasets/Fancylalala/AEGIS}} and the Who\&When~\cite{zhang2025agent} datasets\footnote{\url{https://huggingface.co/datasets/Kevin355/Who_and_When}}. AEGIS-Bench serves as the in-domain dataset and has pre-defined train/val/test splits. On the other hand,  Who\&When is utilized as the out-of-distribution datasets and has exactly one faulty agent per conversation. During training phase, the training and validation splits of AEGIS-Bench are used to train and validate \method; the test split of AEGIS-Bench and the entire Who\&When are used to report the evaluation performances. The data statistics is in Appendix~\ref{app:data-stat}.

\paragraph{Baselines.}


We primarily compare \method\ against a broad range of LLM-based methods, including pre-trained, fine-tuned and proprietary LLMs. Following \citet{kong2025aegis}, we adopt the \textit{All-at-Once} prompting strategy for all LLM baselines, where the model is provided with the user query together with the complete failure trajectory, and is directly asked to identify the faulty agent as well as the corresponding error type \textit{all at once}. The evaluated LLM backbones include \texttt{Qwen2.5}~\cite{qwen2.5}, \texttt{Qwen3}~\cite{yang2025qwen3}, \texttt{GPT-4.1}~\cite{openai2025gpt41}, \texttt{GPT-4o-mini}~\cite{openai2025gpt4omini}, \texttt{o3}~\cite{openai2025gpt4o-o3}, \texttt{Gemini-2.5-Flash/Pro}~\cite{comanici2025gemini} and \texttt{Claude-Sonnet-4}~\cite{anthropic2025system}. We also consider the supervised fine-tuning (\textbf{+S}) and subsequent GRPO-trained (\textbf{+S+G}) variants on small-scale \texttt{Qwen2.5} and \texttt{Qwen3} models.












\paragraph{Evaluation Metrics.}
Following \citet{kong2025aegis}, we evaluate the attribution accuracy at three levels of granularity: \textbf{Pair-level} (correct agent-error pairs), \textbf{Agent-level} (correct faulty agents, ignoring the error types) and \textbf{Error-level} (correct error types, ignoring faulty agents). For each level, \textbf{Micro-F1} ($\mu$F1) and \textbf{Macro-F1} (MF1) are reported. Micro-F1 pools predictions over all samples to compute a single global score. In contrast, Macro-F1 calculates the F1 score separately for each class (e.g., each of the error types) and then averages them uniformly. The details of the error type taxonomy is in Appendix~\ref{app:error-type}.

\paragraph{Implementation Details.}
For \method, we default its backbone GNN to a 2-layer GCN~\cite{kipf2016semi} with a hidden dimension of 64 and a bottleneck dimension of 32. The model is trained using the Adam optimizer with a learning rate of $1\times 10^{-2}$ for up to 100 epochs, with early stopping based on validation performance using a patience of 10 epochs. We set the batch size to 2000 and apply dropout rate 0.1. We employ \texttt{all-MiniLM-L6-v2}~\cite{reimers-2019-sentence-bert} as the embedding model to encode turn node contexts. The best model is selected based on the validation pair-level Micro-F1 score. For \method\ inference, we utilize Micro-F1 to select agent threshold $\tau$ and error threshold $\tau_{fm}$ on AEGIS-Bench; we utilize ranking-based decoding on Who\&When to capture the single error nature. All experiments for \method\ are conducted on an NVIDIA V100 GPU. For the LLM-based methods, we fine-tune \texttt{Qwen2.5-7B-Instruct} \& \texttt{Qwen2.5-14B-Instruct} ourselves  and evaluate them under the same setting. The versions we re-implement are marked with $^{**}$ in Table~\ref{tab:main_tab}. The prompts and the evaluation scripts for the re-implementation are from the AEGIS~\cite{kong2025aegis} codebase\footnote{\url{https://github.com/kfq20/AEGIS}}.

    

\subsection{Main Results}































\begin{table*}[t]
\centering
\small
\setlength{\tabcolsep}{5pt}
\renewcommand{\arraystretch}{1.15}
\caption{Efficiency comparison between \method\ and LLM-based baselines. For \method, preprocessing time summed over  train/validation/test graph construction time. Inference time is reported as in-domain / OOD, respectively.}
\begin{tabular}{lcccc}
\toprule
\midrule

\multicolumn{1}{c}{\textbf{Method} }
& \textbf{Training Time} 
& \textbf{Inference Time} 
& \textbf{Preprocessing Time} 
& \textbf{Trainable Params} \\

\midrule

 7B SFT
&  6 hrs
&  199s / 108s
&  --
&  7B \\

 7B SFT + GRPO
&  > 26 hrs
&  199s / 108s
&  --
&  7B \\

\midrule

 14B SFT
&  8.8 hrs
&  367s / 231s
&  --
&  14B \\

 14B SFT + GRPO
&  > 74 hrs
&  367s / 231s
&  --
&  14B \\

\midrule
  \method
&   1.1 hrs
&   1.16s / 0.37s
&   80.8s
&   65K \\

\midrule
\bottomrule

\end{tabular}

\label{tab:efficiency}
\end{table*}

\paragraph{\method\ is comparable to or outperforms LLMs.}

We present our main results in Table~\ref{tab:main_tab}. \method\  achieves competitive or superior average performance compared to pretrained, post-trained and proprietary LLM baselines over two in-domain and OOD datasets. In particular, despite not relying on LLM-based reasoning, \method\ is highly effective w.r.t. the pair-level metrics. 

We note that pair-level attribution is the most challenging and practically important, since it requires simultaneously identifying both the faulty agent and the correct error type. On AEGIS-Bench, \method\ achieves top-1 pair-level performances. More importantly, under the OOD setting on Who\&When, \method\ remains highly-ranked on pair-level attribution, demonstrating strong robustness under distribution shift. We hypothesize that this advantage mainly comes from modeling structural interaction dynamics and consistency patterns, which remain relatively stable across trajectories; however, post-trained approaches may overfit to surface-level semantic patterns or dataset-specific failure distributions.



\paragraph{\method\ is much more efficient than LLMs.}

As shown in Figure~\ref{fig:overall-comparison} and Table~\ref{tab:efficiency}, \method\ achieves significantly lower training time, inference time and trainable parameters compared to LLM counterpart. The preprocessing time of \method\ is also neglectable (i.e. the time for conversation graph construction over all splits).


\subsection{Discussions}

\paragraph{Backbone sensitivity.}

\begin{table*}[t]
\centering
\setlength{\tabcolsep}{4pt}
\renewcommand{\arraystretch}{1.1}
\caption{Backbone sensitivity study of \method. Best and second-best results are in \textbf{bold} and \underline{underline}. We show that \method\ remains robust across different GNN architectures and model depths. All scores are percentages (\%). The reference model uses a 2-layer GCN with both edge types.}
\resizebox{\textwidth}{!}{
\begin{tabular}{lccccccccccccc}
\toprule
\midrule

\multicolumn{1}{c}{\multirow{4}{*}{\textbf{Backbone}} }
& \multicolumn{6}{c}{\textbf{AEGIS-Bench}} 
& \multicolumn{6}{c}{\textbf{Who\&When}}
& \multirow{4}{*}{\textbf{Avg.}} \\

\cmidrule(lr){2-7}
\cmidrule(lr){8-13}

& \multicolumn{2}{c}{\textbf{Agent}}
& \multicolumn{2}{c}{\textbf{Error}}
& \multicolumn{2}{c}{\textbf{Pair}}
& \multicolumn{2}{c}{\textbf{Agent}}
& \multicolumn{2}{c}{\textbf{Error}}
& \multicolumn{2}{c}{\textbf{Pair}} \\

\cmidrule(lr){2-3}
\cmidrule(lr){4-5}
\cmidrule(lr){6-7}
\cmidrule(lr){8-9}
\cmidrule(lr){10-11}
\cmidrule(lr){12-13}

& $\mu$F1 & MF1 & $\mu$F1 & MF1 & $\mu$F1 & MF1
& $\mu$F1 & MF1 & $\mu$F1 & MF1 & $\mu$F1 & MF1 & \\

\midrule

& \multicolumn{13}{c}{\textbf{Main Model}} \\
\midrule 

GCN (2-layer, all edges)
& \underline{74.16} & \underline{47.86} & \textbf{27.01} & \textbf{25.96} & \underline{17.42} & \textbf{16.35}
& \textbf{37.93} & 20.26 & \textbf{13.79} & \textbf{6.05} & \textbf{6.90}& \textbf{4.16}
& \textbf{24.82} \\
\midrule 
& \multicolumn{13}{c}{\textbf{GNN Backbone}}\\
\midrule
GAT (2-layer, all edges)
& 71.92 & 43.31 & 24.23 & 23.13 & 16.57 & 13.38
& \underline{34.48} & \textbf{25.57} & 7.47 & 3.67 & 0.46 & 0.25
& 22.04 \\

GraphSAGE (2-layer, all edges)
& \textbf{74.34} & \textbf{50.76} & 24.94 & \underline{24.68} & 16.42 & 14.30
& 30.46 & 23.69 & 8.05 & 4.53 & \underline{3.45} & 1.55
& 23.10 \\
\midrule 
& \multicolumn{13}{c}{\textbf{GNN Depth}}\\
\midrule 
GCN (1-layer, all edges)
& 71.17 & 42.17 & 23.23 & 21.88 & 16.46 & \underline{14.39}
& \underline{34.48} & \underline{24.78} & 5.17 & 3.17 & 1.15 & 0.85
& 21.58 \\

GCN (3-layer, all edges)
& 72.84 & 44.25 & \underline{25.05} & 24.61 & \textbf{17.96} & 14.22
& 31.61 & 20.45 & \underline{9.77} & \underline{5.62} & \underline{3.45} & \underline{1.85}
& \underline{22.64} \\

\midrule
\bottomrule
\end{tabular}
}
\label{tab:backbone}
\end{table*}


We present backbone sensitivity in Table~\ref{tab:ablation} to evaluate the robustness of \method\ under different architectural choices.  We observe that the overall performance remains relatively stable across different GNN architectures and layer configurations. 
These observations suggest that the effectiveness of \method\ is not tied to a specific GNN implementation or carefully tuned depth, indicating good architectural robustness.

\paragraph{Ablation.}
\begin{table*}[t]
\centering
\setlength{\tabcolsep}{4pt}
\renewcommand{\arraystretch}{1.1}
\caption{Ablation study of \method. Best and second-best results are in \textbf{bold} and \underline{underline}. We show that graph structure, edge design and each feature component all contribute to \method's performance. All scores are percentages (\%). The reference model uses a 2-layer GCN with both edge types.}
\resizebox{\textwidth}{!}{
\begin{tabular}{lccccccccccccc}
\toprule
\midrule

\multicolumn{1}{c}{\multirow{4}{*}{\textbf{Setting}} }
& \multicolumn{6}{c}{\textbf{AEGIS-Bench}} 
& \multicolumn{6}{c}{\textbf{Who\&When}}
& \multirow{4}{*}{\textbf{Avg.}} \\

\cmidrule(lr){2-7}
\cmidrule(lr){8-13}

& \multicolumn{2}{c}{\textbf{Agent}}
& \multicolumn{2}{c}{\textbf{Error}}
& \multicolumn{2}{c}{\textbf{Pair}}
& \multicolumn{2}{c}{\textbf{Agent}}
& \multicolumn{2}{c}{\textbf{Error}}
& \multicolumn{2}{c}{\textbf{Pair}} \\

\cmidrule(lr){2-3}
\cmidrule(lr){4-5}
\cmidrule(lr){6-7}
\cmidrule(lr){8-9}
\cmidrule(lr){10-11}
\cmidrule(lr){12-13}

& $\mu$F1 & MF1 & $\mu$F1 & MF1 & $\mu$F1 & MF1
& $\mu$F1 & MF1 & $\mu$F1 & MF1 & $\mu$F1 & MF1 & \\

\midrule
& \multicolumn{13}{c}{\textbf{Main Model}} \\
\midrule
GCN (all edges)
& \underline{74.16} & 47.86 & \textbf{27.01} & \textbf{25.96} & 17.42 & \underline{16.35}
& \textbf{37.93} & 20.26 & \textbf{13.79} & \textbf{6.05} & \textbf{6.90} & \textbf{4.16}
& \textbf{24.82} \\

\midrule

& \multicolumn{13}{c}{\textbf{Edge Type Ablation}} \\
\midrule
GCN (no edges)
& 71.26 & 43.21 & 24.65 & 23.54 & \textbf{18.49} & 15.03
& 32.18 & 25.10 & 8.62 & 5.55 & 3.45 & 2.39
& 22.78 \\

GCN (only same-agent edges)
& 70.53 & 42.74 & 22.32 & 21.58 & 16.31 & 12.23
& \underline{36.78} & \underline{26.21} & 6.32 & 2.95 & \underline{4.60} & \underline{3.16}
& 22.14 \\

GCN (only temporal edges)
& \textbf{74.82} & \textbf{53.60} & 24.82 & 24.39 & 17.54 & \textbf{16.52}
& 35.06 & 20.41 & 5.75 & 3.41 & \underline{4.60} & 1.99
& \underline{23.58} \\
\midrule
& \multicolumn{13}{c}{\textbf{Model Design Ablation}} \\
\midrule
W/o dev/stat features
& 65.90 & 35.47 & 23.59 & 22.25 & 14.00 & 9.80
& \textbf{37.93} & \textbf{32.00} & 5.75 & 2.89 & 1.72 & 1.14
& 21.04 \\

W/o sentence embedding
& 73.58 & \underline{48.18} & 23.62 & 22.14 & 14.78 & 10.16
& \textbf{37.93} & 21.76 & 8.05 & 5.44 & 2.30 & 1.33
& 22.44 \\

W/o GNN
& 71.11 & 43.80 & \underline{26.34} & \underline{25.61} & \underline{17.81} & 14.93
& 26.44 & 18.86 & \underline{11.49} & \underline{5.68} & 2.87 & 1.89
& 22.24 \\

\midrule
\bottomrule
\end{tabular}
}

\label{tab:ablation}
\end{table*}

In Table~\ref{tab:ablation}, we further study the impact of connection types and some key model designs. Firstly, removing all edges consistently degrades performance, demonstrating that relational structure is important for agent failure attribution. Secondly, using only same-agent edges or only temporal edges also leads to noticeable performance drops compared to the full graph, suggesting that both temporal interaction patterns and intra-agent behavioral consistency provide complementary signals. 

For model design ablation, we observe that removing deviation/statistical features causes large overall performance degradation, particularly on pair-level metrics, supporting our hypothesis that faulty agents are often characterized by abnormal interaction dynamics and consistency violations. Removing the GNN module also harms performance, showing the importance of explicit structured modeling.

\paragraph{Generalization study.}
\begin{table}[t]
\centering
\setlength{\tabcolsep}{4pt}
\renewcommand{\arraystretch}{1.1}
\caption{Test-time adaptation results on Who\&When. Best and second-best results are in \textbf{bold} and \underline{underline}.  We observe that pair-level adaptation achieves the strongest overall gains.}
\resizebox{0.8\textwidth}{!}{
\begin{tabular}{clccccccc}
\toprule
\midrule

\multirow{2}{*}{\textbf{Category}} 
& \multicolumn{1}{c}{\multirow{2}{*}{\textbf{Method}}}
& \multicolumn{2}{c}{\textbf{Agent}}
& \multicolumn{2}{c}{\textbf{Error}}
& \multicolumn{2}{c}{\textbf{Pair}}
& \multirow{2}{*}{\textbf{Avg.}} \\

\cmidrule(lr){3-4}
\cmidrule(lr){5-6}
\cmidrule(lr){7-8}

&
& $\mu$F1 & MF1
& $\mu$F1 & MF1
& $\mu$F1 & MF1
& \\

\midrule

\multirow{1}{*}{\textbf{Original}}
& Base
& \underline{38.51} & \underline{22.37}
& 9.77 & 5.05
& 5.17 & 2.92
& 13.96 \\

\midrule

\multirow{3}{*}{\textbf{TTA}}
& Agent
& 35.06 & 21.42
& \textbf{14.37} & \underline{5.56}
& \underline{6.32} & \underline{4.40}
& 14.52 \\

& Pair
& 35.06 & 20.15
& \textbf{14.37} & \textbf{5.77}
& \textbf{7.47} & \textbf{4.96}
& \underline{14.63} \\

& Pair-Faulty
& \textbf{39.08} & \textbf{22.47}
& 11.49 & 5.06
& \underline{6.32} & 4.06
& \textbf{14.75} \\

\midrule
\bottomrule
\end{tabular}
}
\label{tab:tta}
\end{table}
While \method\ demonstrates strong performance on the in-domain benchmark, generalizing agent failure attribution under severe distribution shift remains an inherently challenging problem. As a case study, we investigate whether lightweight test-time adaptation (TTA) can improve \method\ under OOD settings without requiring retraining or additional supervision.

Motivated by entropy-minimization (EM) based test-time adaptation such as TENT~\cite{wang2020tent}, we explore whether \method\ can adapt to OOD trajectories by encouraging more confident predictions at inference time. We consider three adaptation objectives at different prediction granularities as follows: (i) \textbf{Agent-level EM} minimizes the binary entropy of agent faulty probability; (ii) \textbf{Pair-level EM} encourages confident predictions over pairwise label space; and (iii) \textbf{Faulty-agent pair-level EM} applies pair-level entropy minimization only to agents that are likely to be faulty. The TTA procedure and the detailed formulas of these strategies are in Appendix~\ref{app:tta}. Table~\ref{tab:tta} shows that  lightweight test-time adaptation can consistently improve OOD performance.

\section{Related Work}


\paragraph{Direct prompting.}

Early approaches to agent failure attribution primarily rely on prompting pretrained LLMs to directly analyze interaction trajectories~\cite{zhang2025agent, deshpande2025trail}. These methods treat attribution as a reasoning task over long execution logs. However, these approaches rely on costly LLM inference and often require carefully designed prompting strategies.

\paragraph{Finetuning-based methods.}

Another line of work focuses on training specialized attribution models using synthetic data with labeled faulty trajectories~\cite{zhang2025graphtracer, zhang2025agentracer, kong2025aegis}. These methods generate large-scale annotated failure trajectories through techniques such as error injection, counterfactual replay, or graph-guided synthesis, and subsequently post-train models using supervised fine-tuning and reinforcement learning. However, these methods introduce significant overhead in training cost and they still inherit the limitations of direct LLM inference at test time. 

\paragraph{Agentic systems.}

Some works explores more complex failure attribution frameworks. In this category, some approaches employ more sophisticate prompting techniques~\cite{zhu2026raffles}, some construct structured representations such as causal graphs~\cite{wang2026flat}, hierarchical context models~\cite{banerjee2025did} and complex patterns~\cite{ge2025introducing, sun2026scope} to better capture inter-agent dependencies, while others incorporate memory mechanisms to reuse previously observed failure patterns~\cite{yu2025correct}.


\section{Conclusion}

We revisit agent failure attribution in LLM-based multi-agent systems and show that heavy generative approaches are not necessary for strong performance. 
We propose \method, a lightweight graph-based framework that models interaction trajectories through simple step-level signals and agent-level structure, achieving competitive or superior results against LLM-based methods with minimal computational cost.  These findings suggest that structured modeling of agent interactions is sufficient for effective attribution. Future work includes designing more tailored aggregation operators  and conducting in-depth analysis of generalization under distribution shifts.

\newpage
\bibliographystyle{plainnat}
\bibliography{Reference}

\newpage
\appendix

\section{Graph construction}

\label{app:node-features}



\paragraph{TF-IDF computation.}
For each conversation, we first construct a local vocabulary from all turns after lowercasing and tokenization. Based on this vocabulary, we compute turn-level TF-IDF representations using standard term frequency and smoothed inverse document frequency, followed by row-wise $\ell_2$ normalization. Let $\mathbf{X}_{\mathrm{TF\mbox{-}IDF}}\in\mathbb{R}^{T\times |\mathcal{V}|}$ denote the resulting TF-IDF matrix, where $T$ is the number of turns and $|\mathcal{V}|$ is the vocabulary size.

To obtain compact dense representations that capture the dominant vocabulary structure within the conversation, we apply truncated SVD:
\[
\mathbf{Z}_{\mathrm{svd}}
=
\operatorname{SVD}_r(\mathbf{X}_{\mathrm{TF\mbox{-}IDF}})
\in
\mathbb{R}^{T\times r},
\]
where $r$ denotes the retained rank size. Each row
\[
\mathbf{z}_{\mathrm{svd}}^t
=
\mathbf{Z}_{\mathrm{svd}}[t,:]
\in
\mathbb{R}^r
\]
provides an $r$-dimensional representation for turn $t$. In our implementation, we use $r=16$.

\paragraph{Deviation features.}
 Let $\phi_t = \mathbf{z}^t_{\mathrm{svd}}$ denote the TF-IDF SVD feature of turn $t$, and let $\mathcal{I}_a=\{t \mid a_{i_t}=a\}$ denote the set of turns generated by agent $a$. 

\begin{itemize}

    \item \textbf{Sequential deviation.}
    We measure how abruptly a turn deviates from the immediately preceding turn:
    \[
    f_{\mathrm{seq}}(t)
    =
    1-\cos(\phi_t,\phi_{t-1}).
    \]

    \item \textbf{Self inconsistency.}
    We measure deviation from the historical behavior of the same agent:
    \[
    \mu_{a_{i_t}}
    =
    \frac{1}{|\mathcal{I}_{a_{i_t}}|}
    \sum_{s\in \mathcal{I}_{a_{i_t}}}\phi_s,
    \]
    \[
    f_{\mathrm{self}}(t)
    =
    1-\cos(\phi_t,\mu_{a_{i_t}}).
    \]

    \item \textbf{Conversation consensus deviation.}
    We compute deviation from the global conversation centroid:
    \[
    \mu_{\mathcal{T}}
    =
    \frac{1}{T}\sum_{s=1}^{T}\phi_s,
    \]
    \[
    f_{\mathrm{cons}}(t)
    =
    1-\cos(\phi_t,\mu_{\mathcal{T}}).
    \]

    \item \textbf{Cross-agent deviation.}
    We measure deviation from the average representation of all other agents:
    \[
    \mu_{\neg a_{i_t}}
    =
    \frac{1}{T-|\mathcal{I}_{a_{i_t}}|}
    \sum_{s\notin \mathcal{I}_{a_{i_t}}}\phi_s,
    \]
    \[
    f_{\mathrm{other}}(t)
    =
    1-\cos(\phi_t,\mu_{\neg a_{i_t}}).
    \]

    \item \textbf{Agent internal consistency.}
    We compute the average pairwise similarity among all turns generated by the same agent:
    \[
    f_{\mathrm{within}}(a)
    =
    \frac{1}{|\mathcal{I}_a|(|\mathcal{I}_a|-1)}
    \sum_{\substack{s,u\in \mathcal{I}_a \\ s\neq u}}
    \cos(\phi_s,\phi_u).
    \]

    \item \textbf{Same-agent temporal consistency.}
    We measure similarity to the previous turn generated by the same agent:
    \[
    p(t)
    =
    \max\{s<t \mid a_{i_s}=a_{i_t}\},
    \]
    \[
    f_{\mathrm{prev}}(t)
    =
    \cos(\phi_t,\phi_{p(t)}).
    \]

    \item \textbf{Vocabulary stability.}
    Let $\mathcal{V}_t$ denote the token set of turn $t$, and let
    \[
    \mathcal{V}_{a_{i_t}}^{(-t)}
    =
    \bigcup_{s\in \mathcal{I}_{a_{i_t}},\,s\neq t}\mathcal{V}_s.
    \]
    We measure the overlap ratio between the current turn and the remaining turns from the same agent:
    \[
    f_{\mathrm{vocab}}(t)
    =
    \frac{
    |\mathcal{V}_t\cap \mathcal{V}_{a_{i_t}}^{(-t)}|
    }{
    |\mathcal{V}_t|
    }.
    \]

    \item \textbf{Problem alignment.}
    We measure similarity between the current turn and the initial problem statement:
    \[
    f_{\mathrm{prob}}(t)
    =
    \cos(\phi_t,\phi_1).
    \]
\end{itemize}

\paragraph{Statistical features.}

\begin{itemize}





    \item \textbf{Numerical statistics.}
    We include lightweight numerical indicators, including whether the turn contains numeric tokens:
    \[
    f_{\mathrm{num}}(t)
    =
    \mathbb{I}[\exists \text{ digit token in } x_t],
    \]
    as well as the normalized mean and standard deviation of word lengths within the conversation:
    \[
    f_{\mathrm{wmean}}(t)
    =
    \frac{
    \operatorname{mean\_len}(x_t)-\mu_{\mathrm{len}}
    }{
    \sigma_{\mathrm{len}}
    },
    \qquad
    f_{\mathrm{wstd}}(t)
    =
    \frac{
    \operatorname{std\_len}(x_t)-\mu_{\mathrm{std}}
    }{
    \sigma_{\mathrm{std}}
    }.
    \]

    \item \textbf{Position statistics.}
    We include normalized turn position
    \[
    f_{\mathrm{pos}}(t)
    =
    \frac{t-1}{T-1},
    \]
    binary indicators for whether the turn is the first or last turn:
    \[
    f_{\mathrm{first}}(t)
    =
    \mathbb{I}[t=1],
    \qquad
    f_{\mathrm{last}}(t)
    =
    \mathbb{I}[t=T],
    \]
    and the cumulative fraction of turns authored by the same agent up to step $t$:
    \[
    f_{\mathrm{agent\_frac}}(t)
    =
    \frac{
    |\{s\le t \mid a_{i_s}=a_{i_t}\}|
    }{
    t
    }.
    \]

    \item \textbf{Length statistics.}
    Let $\ell_t$ denote the token length of turn $t$. We include normalized turn length:
    \[
    f_{\mathrm{len}}(t)
    =
    \frac{\ell_t}{\max_{s}\ell_s},
    \]
    together with the conversation-level z-normalized length:
    \[
    f_{\mathrm{zlen}}(t)
    =
    \frac{\ell_t-\mu_{\ell}}{\sigma_{\ell}}.
    \]

    \item \textbf{Structural statistics.}
    We additionally include the total participation ratio of the corresponding agent:
    \[
    f_{\mathrm{part}}(t)
    =
    \frac{|\mathcal{I}_{a_{i_t}}|}{T},
    \]
    and the spread of the agent's participation positions:
    \[
    f_{\mathrm{spread}}(a)
    =
    \operatorname{std}
    \left(
    \left\{
    \frac{s-1}{T-1}
    \mid s\in \mathcal{I}_a
    \right\}
    \right).
    \]

\end{itemize}


\section{Dataset Details}
\label{app:data-stat}

Both datasets ae under MIT licenses.
\begin{table}[h]
\centering

\begin{tabular}{cccc}
\toprule
\midrule

\multicolumn{3}{c}{\textbf{AEGIS-Bench}~\cite{kong2025aegis}} 
& \textbf{Who\&When}~\cite{zhang2025agent} \\

\cmidrule(lr){1-3}
\cmidrule(lr){4-4}

\textbf{Train} & \textbf{Validation} & \textbf{Test} & \textbf{Test} \\

\midrule

7146 & 1787 & 600 & 184 \\

\midrule
\bottomrule

\end{tabular}
\vspace{2mm}
\caption{Dataset size statistics used in our experiments.}
\label{tab:data_stats}

\end{table}

\section{Error Types}
\label{app:error-type}
Adopted from~\cite{kong2025aegis}, we consider the following 14 error types in agent failure attribution:

\begin{itemize}

    \item \textbf{Specification Issues}
    \begin{itemize}
        \item FM-1.1: Task specification deviation
        \item FM-1.2: Role specification deviation
        \item FM-1.3: Add redundant steps
        \item FM-1.4: Remove conversation history
        \item FM-1.5: Remove termination conditions
    \end{itemize}

    \item \textbf{Inter-Agent Misalignment}
    \begin{itemize}
        \item FM-2.1: Repeat handled tasks
        \item FM-2.2: Make request ambiguous
        \item FM-2.3: Deviate from main goal
        \item FM-2.4: Hide important information
        \item FM-2.5: Ignore other agents
        \item FM-2.6: Inconsistent reasoning
    \end{itemize}

    \item \textbf{Task Verification Failures}
    \begin{itemize}
        \item FM-3.1: Premature termination
        \item FM-3.2: Remove verification steps
        \item FM-3.3: Incorrect verification
    \end{itemize}

\end{itemize}



\section{Test-Time Adaptation for \method}
\label{app:tta}

Below we detail the loss objective for each TTA strategy.

\paragraph{Agent-level entropy minimization.}
Let $a_i$ denote the scalar faulty-agent logit for agent $i$, and let $p_i=\sigma(a_i)$. We define
\[
\mathcal{L}_{\mathrm{agent}}
=
\frac{1}{N_A}
\sum_{i=1}^{N_A}
H_{\mathrm{bin}}(p_i)
=
-\frac{1}{N_A}
\sum_{i=1}^{N_A}
\left[
p_i \log p_i
+
(1-p_i)\log(1-p_i)
\right].
\]

\paragraph{Pair-level entropy minimization.}

Let $\mathbf{s}_i\in\mathbb{R}^{K+1}$ denote the pair logits for agent $i$, where the first class corresponds to the clean class and the remaining $K$ classes correspond to error types. With $\hat{p}_{ic}=\mathrm{softmax}(\mathbf{s}_i)_c$, we minimize
\[
\mathcal{L}_{\mathrm{pair}}
=
\frac{1}{N_A}
\sum_{i=1}^{N_A}
H_{\mathrm{cat}}(\mathbf{s}_i)
=
-\frac{1}{N_A}
\sum_{i=1}^{N_A}
\sum_{c=0}^{K}
\hat{p}_{ic}\log \hat{p}_{ic}.
\]

\paragraph{Faulty-agent pair-level minimization.}

We first estimate the number of faulty agents as
$k
=
\max\left(
1,
\sum_{i=1}^{N_A}
\mathbb{I}[a_i>0]
\right)$ 
and select the top-$k$ agents according to the faulty-agent logits $a_i$. The adaptation objective is then
\[
\mathcal{L}_{\mathrm{pair\text{-}faulty}}
=
\frac{1}{k}
\sum_{i\in \mathrm{Top}\text{-}k(a)}
H_{\mathrm{cat}}(\mathbf{s}_i)
=
-\frac{1}{k}
\sum_{i\in \mathrm{Top}\text{-}k(a)}
\sum_{c=0}^{K}
\hat{p}_{ic}\log \hat{p}_{ic}.
\]

For the TTA procedure, we freeze all parameters except the LayerNorm affine parameters across all layers in \method\ and minimize the above three losses. We use Adam optimizer with learning rate $1e-3$ for 30 gradient steps over the entire OOD batch. No OOD labels are used. The final checkpoint is taken as the final model for evaluation.

\section{Limitation}
\label{app:limitation}

\begin{itemize}

    \item \textbf{OOD generalization.}
    Although \method\ demonstrates certain robustness under distribution shift, generalization across substantially different multi-agent settings remains challenging. Future work may include designing customized message passing operators and adaptive graph propagation mechanisms that better capture failure propagation patterns.

    \item \textbf{Limited reasoning capability.}
    While lightweight structural modeling is effective for our studied attribution scenarios, certain failures may still require deeper semantic understanding and long-horizon reasoning. An important future direction is to combine graph-based modeling with LLM  reasoning to jointly leverage structural consistency and semantic inference.

\end{itemize}

\section{Impact Statement}
\label{app:impact}
This paper discusses the advancement of the field of Large Language Model and Graph Machine Learning. While there are potential societal consequence of our work, none of which we feel must be hightlighted.




\newpage
\section*{NeurIPS Paper Checklist}

\begin{enumerate}

\item {\bf Claims}
    \item[] Question: Do the main claims made in the abstract and introduction accurately reflect the paper's contributions and scope?
    \item[] Answer: \answerYes{} 
    \item[] Justification: Our main contributions are summarized as four bullet points in Section 1 and each of them are supported by claims in introduction or experiment section.

\item {\bf Limitations}
    \item[] Question: Does the paper discuss the limitations of the work performed by the authors?
    \item[] Answer: \answerYes{} 
    \item[] Justification: We discuss potential limitations of our presented work in Appendix~\ref{app:limitation}.

\item {\bf Theory assumptions and proofs}
    \item[] Question: For each theoretical result, does the paper provide the full set of assumptions and a complete (and correct) proof?
    \item[] Answer: \answerNA{} 
    \item[] Justification: We don't include theoretical results.

    \item {\bf Experimental result reproducibility}
    \item[] Question: Does the paper fully disclose all the information needed to reproduce the main experimental results of the paper to the extent that it affects the main claims and/or conclusions of the paper (regardless of whether the code and data are provided or not)?
    \item[] Answer: \answerYes{} 
    \item[] Justification: We provide all experimental details needed in Section~\ref{ssec:exp-setup}.

\item {\bf Open access to data and code}
    \item[] Question: Does the paper provide open access to the data and code, with sufficient instructions to faithfully reproduce the main experimental results, as described in supplemental material?
    \item[] Answer: \answerYes{} 
    \item[] Justification: All the benchmarks used in this work is public. We will provide the code package during submission and make the code
available upon acceptance.

\item {\bf Experimental setting/details}
    \item[] Question: Does the paper specify all the training and test details (e.g., data splits, hyperparameters, how they were chosen, type of optimizer) necessary to understand the results?
    \item[] Answer: \answerYes{} 
    \item[] Justification: We provide all experimental details needed in Section~\ref{ssec:exp-setup}.

\item {\bf Experiment statistical significance}
    \item[] Question: Does the paper report error bars suitably and correctly defined or other appropriate information about the statistical significance of the experiments?
    \item[] Answer: \answerNo{} 
    \item[] Justification: Our conclusions are well supported by consistent trends across multiple datasets, evaluation metrics, model backbones and ablation settings.

\item {\bf Experiments compute resources}
    \item[] Question: For each experiment, does the paper provide sufficient information on the computer resources (type of compute workers, memory, time of execution) needed to reproduce the experiments?
    \item[] Answer: \answerYes{} 
    \item[] Justification:  We provide all experimental details needed in Section~\ref{ssec:exp-setup}.
    
\item {\bf Code of ethics}
    \item[] Question: Does the research conducted in the paper conform, in every respect, with the NeurIPS Code of Ethics \url{https://neurips.cc/public/EthicsGuidelines}?
    \item[] Answer: \answerYes{} 
    \item[] Justification: We confirm that this work is conducted with the NeurIPS Code of Ethics.

\item {\bf Broader impacts}
    \item[] Question: Does the paper discuss both potential positive societal impacts and negative societal impacts of the work performed?
    \item[] Answer: \answerYes{} 
    \item[] Justification: We provide related discussion in Appendix~\ref{app:impact}.    \item[] Guidelines:
    \begin{itemize}
        \item The answer \answerYes{} means that there is no societal impact of the work performed.
        \item If the authors answer \answerNA{} or \answerNo, they should explain why their work has no societal impact or why the paper does not address societal impact.
        \item Examples of negative societal impacts include potential malicious or unintended uses (e.g., disinformation, generating fake profiles, surveillance), fairness considerations (e.g., deployment of technologies that could make decisions that unfairly impact specific groups), privacy considerations, and security considerations.
        \item The conference expects that many papers will be foundational research and not tied to particular applications, let alone deployments. However, if there is a direct path to any negative applications, the authors should point it out. For example, it is legitimate to point out that an improvement in the quality of generative models could be used to generate Deepfakes for disinformation. On the other hand, it is not needed to point out that a generic algorithm for optimizing neural networks could enable people to train models that generate Deepfakes faster.
        \item The authors should consider possible harms that could arise when the technology is being used as intended and functioning correctly, harms that could arise when the technology is being used as intended but gives incorrect results, and harms following from (intentional or unintentional) misuse of the technology.
        \item If there are negative societal impacts, the authors could also discuss possible mitigation strategies (e.g., gated release of models, providing defenses in addition to attacks, mechanisms for monitoring misuse, mechanisms to monitor how a system learns from feedback over time, improving the efficiency and accessibility of ML).
    \end{itemize}
    
\item {\bf Safeguards}
    \item[] Question: Does the paper describe safeguards that have been put in place for responsible release of data or models that have a high risk for misuse (e.g., pre-trained language models, image generators, or scraped datasets)?
    \item[] Answer: \answerNA{} 
    \item[] Justification: We confirm that this work does not pose safety risks.

\item {\bf Licenses for existing assets}
    \item[] Question: Are the creators or original owners of assets (e.g., code, data, models), used in the paper, properly credited and are the license and terms of use explicitly mentioned and properly respected?
    \item[] Answer: \answerYes{} 
    \item[] Justification: We detail them in Appendix~\ref{app:data-stat}.

\item {\bf New assets}
    \item[] Question: Are new assets introduced in the paper well documented and is the documentation provided alongside the assets?
    \item[] Answer: \answerNA{} 
    \item[] Justification: No new assets introduced.

\item {\bf Crowdsourcing and research with human subjects}
    \item[] Question: For crowdsourcing experiments and research with human subjects, does the paper include the full text of instructions given to participants and screenshots, if applicable, as well as details about compensation (if any)? 
    \item[] Answer: \answerNA{} 
    \item[] Justification: This paper does not involve human subjects.

\item {\bf Institutional review board (IRB) approvals or equivalent for research with human subjects}
    \item[] Question: Does the paper describe potential risks incurred by study participants, whether such risks were disclosed to the subjects, and whether Institutional Review Board (IRB) approvals (or an equivalent approval/review based on the requirements of your country or institution) were obtained?
    \item[] Answer: \answerNA{} 
    \item[] Justification: This paper does not involve human subjects.

\item {\bf Declaration of LLM usage}
    \item[] Question: Does the paper describe the usage of LLMs if it is an important, original, or non-standard component of the core methods in this research? Note that if the LLM is used only for writing, editing, or formatting purposes and does \emph{not} impact the core methodology, scientific rigor, or originality of the research, declaration is not required.
    \item[] Answer: \answerNA{} 
    \item[] Justification: The proposed method is entirely LLM-free and does not use LLMs as part of its core methodology. Although some comparison baselines are LLM-based methods, they are only included for evaluation purposes and are not key components of the proposed framework.

\end{enumerate}

\end{document}